\documentclass[conference]{IEEEtran}
\IEEEoverridecommandlockouts
\usepackage{cite}
\usepackage{amsmath,amssymb,amsfonts}
\usepackage{algorithmic}
\usepackage{graphicx}
\usepackage{textcomp}
\usepackage{xcolor}
\usepackage{float}
\usepackage{microtype}
\def\BibTeX{{\rm B\kern-.05em{\sc i\kern-.025em b}\kern-.08em
    T\kern-.1667em\lower.7ex\hbox{E}\kern-.125emX}}
\begin{document}

\title{Automated Immunophenotyping Assessment for Diagnosing Childhood Acute Leukemia using Set-Transformers\\
\thanks{This project has received funding from the European Union’s Horizon 2020 research and innovation programme under the Marie Skłodowska-Curie grant agreement No 101034277.}
}

\author{
    \IEEEauthorblockN{Elpiniki Maria Lygizou\IEEEauthorrefmark{1},
    Michael Reiter\IEEEauthorrefmark{1}, 
    Margarita Maurer-Granofszky\IEEEauthorrefmark{2}, Michael Dworzak\IEEEauthorrefmark{2}, Radu Grosu\IEEEauthorrefmark{1}}
    
    \IEEEauthorblockA{\IEEEauthorrefmark{1}TU Wien\\
    elpiniki.lygizou@tuwien.ac.at, rei@cvl.tuwien.ac.at, radu.grosu@tuwien.ac.at}
    
    \IEEEauthorblockA{\IEEEauthorrefmark{2}St. Anna Children's Cancer Research Institute\\
    margarita.maurer@ccri.at, michael.dworzak@ccri.at}
}


\maketitle

\begin{abstract}
Acute Leukemia is the most common hematologic malignancy in children and adolescents. A key methodology in the diagnostic evaluation of this malignancy is immunophenotyping based on Multiparameter Flow Cytometry (FCM). However, this approach is manual, and thus time-consuming and subjective. To alleviate this situation, we propose in this paper the FCM-Former, a machine learning, self-attention based FCM-diagnostic tool, automating the immunophenotyping assessment in Childhood Acute Leukemia. The FCM-Former is trained in a supervised manner, by directly using flow cytometric data. Our FCM-Former achieves an accuracy of $\textbf{96.5\%}$ assigning lineage to each sample among 960 cases of either acute B-cell, T-cell lymphoblastic, and acute myeloid leukemia (B-ALL, T-ALL, AML). To the best of our knowledge, the FCM-Former is the first work that automates the immunophenotyping assessment with FCM data in diagnosing pediatric Acute Leukemia.
\end{abstract}

\begin{IEEEkeywords}
immunophenotyping, multiparameter flow cytometry, set-transformers, self-attention 
\end{IEEEkeywords}

\section{Introduction}
Acute Leukemias are a heterogeneous group of hematologic malignancies  (cancers), which progress rapidly. Hence, their prompt detection is crucial for a successful treatment.
These diseases are primarily categorized based on the lineage of the affected cells, referring to the type of precursor cells (early-form cells), that begin to multiply at an accelerated rate.

Immunophenotyping is an essential part of the precise diagnosis and classification of acute leukemia \cite{c5}, although it is manual, time-consuming, and subjective, by relying on the experience and knowledge of domain experts. A reliable tool for immunophenotyping is flow cytometry, a laser-based biophysical technique which provides a quick and comprehensive multi-parameter analysis of individual cells or particles.

In order to automate the immunophenotyping assessment in the diagnosis of Childhood Acute Leukemia, we introduce in this paper the FCM-Former, a machine learning and self-attention-based classification algorithm 
for FCM data. The FCM-Former is based on the Set-Transformer architecture of~\cite{c2}, and it is designed to work directly with the FCM data obtained from CCRI in Vienna, without any additional pre-processing. This direct approach, also allows the FCM-Former to take advantage of the high dimensionality of the FCM data. The main goal of the FCM-Former, is to accurately classify the malignancy into one of three lineages: B-ALL, T-ALL and AML. To the best of our knowledge, the FCM-Former is the first work that automates the immunophenotyping assessment with FCM data in diagnosing pediatric Acute Leukemia. Our work thus represents an important step towards applying advanced machine  learning techniques, to critical healthcare challenges. In particular, in the accurate and timely diagnosis of pediatric acute leukemia.

The rest of this paper is structured as follows. 
In Section~2 we review foundational concepts for this work. In Section~3, we discuss the related work and previous approaches. In Section~4, we describe the experimental methodology, leading to Section~5, where we present our experimental results. Finally, we conclude in Section~6, by summarizing our key insights, and proposing directions for future work.

\section{Background}

\subsection{Multiparameter Flow Cytometry}

Multiparameter flow cytometry (FCM) serves as a robust and powerful tool for both analytical and preparative applications \cite{c21},\cite{c22}. In FCM, a blood or bone-marrow sample of patients is stained with a specific combination of fluorochrome-labelled antibodies (markers) uniquely binding to antigens, intracellular or on cell's surface. The resulting data are a set of measurements (feature vectors) of the physical (size, granularity), and biological (multiple surface/intracellular markers) properties of every single cell (an event). This set (sample) thus characterizes the phenotype of a hole cell population. 

\subsection{Transformers}

In this section, we delve into the core principles of the Transformer~\cite{c1} and Set-Transformer~\cite{c2} models, focusing on the self-attention mechanisms behind them.

\vspace*{1mm}\textbf{Transformers.} These are advanced deep learning models, primarily developed for natural language processing (NLP). Their unique architecture, is characterized by a self-attention mechanism, allowing them to focus on complex relationships within data, and capture meaningful patterns in large-scale data. This leads to an effective context understanding.
The multi-headed self-attention mechanism, as introduced in \cite{c1}, is defined for a given set of $n$ query vectors $Q$ ($n$ corresponds to the set's size consists of $n$ elements) each with dimension $d_q$, $Q\,{\in}\,\mathbb{R}^{n_q \times d_q}$, key matrices $K\,{\in}\,\mathbb{R}^{n_v \times d_q}$ and value matrices $V\,{\in}\,\mathbb{R}^{n_v \times d_v}$, where $d_q\,{=}\,d_v\,{=}\,d$ for the sake of simplicity. It can be described as a function according to the formula below:
$$
attn(Q,K,V) := softmax(\frac{QK^T}{\sqrt{d}})V \eqno{(1)}
$$
If $Q$ and $K$ are derived from the same set of inputs (as in self-attention), the $QK^T$ multiplication is quadratic in the set size, which prevents the direct application to FCM data.

\vspace*{1mm}\textbf{Set-transformers (ST).} This derivative of the original transformer architecture, is designed to operate on set-structured data, where ordering is irrelevant to the input information. The associated model adapts the transformer's architecture to handle input data, that lacks a clear sequential or grid-like structure, similar to how the FCM data are in our application.

The building block of Set Transformers reduces the \(O(n^2)\) complexity of self-attention to \(O(nm)\) by incorporating inducing points into the standard multi-head self-attention block of Formula~(1), where $n$ is the input dimension and $m$ is the number of learnable parameters (inducing points).

The process begins by projecting $Q, K, V$ onto $h$ different $d_h^q, d_h^q, d_h^v,$-dimensional vectors, where $d_h^q\,{=}\,d_h^v\,{=}\,\frac{d}{h}$, such that:
$$
Multihead(Q, K, V ) := concat(O_1,..., O_h)W^O \eqno{(2)}
$$
$$\text{where } O_j = attn(QW_j^Q, KW_j^K, VW_j^V) \eqno{(3)}$$
where $W_j^Q, W_j^K, W_j^V$ are projection operators of dimensions $\mathbb{R}^{d_q \times d_h^q}$, $\mathbb{R}^{d_q \times d_h^q}$ and $\mathbb{R}^{d_v \times d_h^v}$, respectively, and $W^O$ is a linear operator of dimension $d \times d$ relating $O_1,{\ldots}\,O_h$ to each other. 

Furthermore, given a set $S$ of $d$-dimension vectors, we initialize $m$ $d$-dimensional inducing points $I\,{\in}\,\mathbb{R}^{m \times d}$. Then, the Multihead Set-Transformer Attention Block (MSAB) is computed by the following formulas:
$$
MSAB(I,S) := LayerNorm(X + rFF(X)) \eqno{(4)}
$$
$$\text{where } X = LayerNorm(I + Multihead(I, S, S)) \eqno{(5)}$$
where rFF denotes a row-wise feedforward layer, and LayerNorm is layer normalization as described in \cite{c6}. Finally, the Set-Transformer Attention Block (STAB) is defined as follows:
$$
STAB(S) := MSAB(S,MSAB(I,S)) \eqno{(6)}
$$

\section{Related Work}

Manual analysis of FCM data, which plays a crucial role in various medical and biological fields, typically involves representing and transforming the high-dimensional space of raw data, into 2-D plots for human interpretation. This technique, while making the data more comprehensible, can lead to a loss of information.
However, machine learning methods, can utilize the full data space, and tackle this shortcoming.  

Automated FCM data analysis, primarily focuses on identifying and classifying distinct or specific cell populations. Initial methods in this domain pooled events from different samples, employing classifiers based on single-event pairs and labels \cite{c9},\cite{c10},\cite{c11}, but were limited to fixed decision regions. This approach was less effective in discerning relational positioning among cell populations, a key factor in detecting rare or abnormal cells, particularly in Minimal Residual Disease (MRD) detection \cite{c12}.
Subsequent developments in FCM analysis therefore shifted towards processing a hole sample in a unified manner. Techniques such as Gaussian Mixture Models (GMM) \cite{c13}, and Convolutional Neural Networks (CNNs) \cite{c14} emerged, which were applied to multiple 2-D projections of the data space. These methods addressed some limitations of earlier approaches, particularly in maintaining relational context among cell populations. They are less suited for tabular data analysis, which characterizes FCM data. In the realm of automated immunophenotyping statistical methods have been proposed employing distance-based analysis in the space of principal components calculated on a database of FCM reference samples \cite{c23}. While these methods rely on strict standardization in the data acquisition process (flow cytometer settings, FCM panels, etc.), our claim is to process data in diverse conditions without the need for such rigid standardization by using machine learning models being able to identify and relate structures in the data space and thus deal to a larger degree with data distortions.

More recently, attention-based models have gained prominence in automated FCM analysis \cite{c3},\cite{c15},\cite{c8},\cite{c16}. These models, using attention mechanisms, emulate the human logic of manual FCM data analysis, but keep and leverage the high-dimension data space information of the FCM data. They enable event-level classification by learning the importance of various cell populations within a sample. They are well known for their SOTA performance in tasks such as automated MRD detection \cite{c3}, and recently, in adults acute leukemia diagnosis \cite{c8}. 

However, these methods often assume a fixed set of features during training and inference, which can be a limitation given the high variability of FCM data features even within a single dataset. Some recent studies have explored combining features from different samples using techniques like nearest neighbor imputation \cite{c18},\cite{c19}, but the efficacy of these approaches is still under scrutiny due to potential inaccuracies in imputed values affecting downstream analysis \cite{c20}. A late work \cite{c16} attempts to address these issues by employing a feature-agnostic, attention-based method with promising results.

\begin{figure*}[!t]
   \centering
   \includegraphics[width=\textwidth]{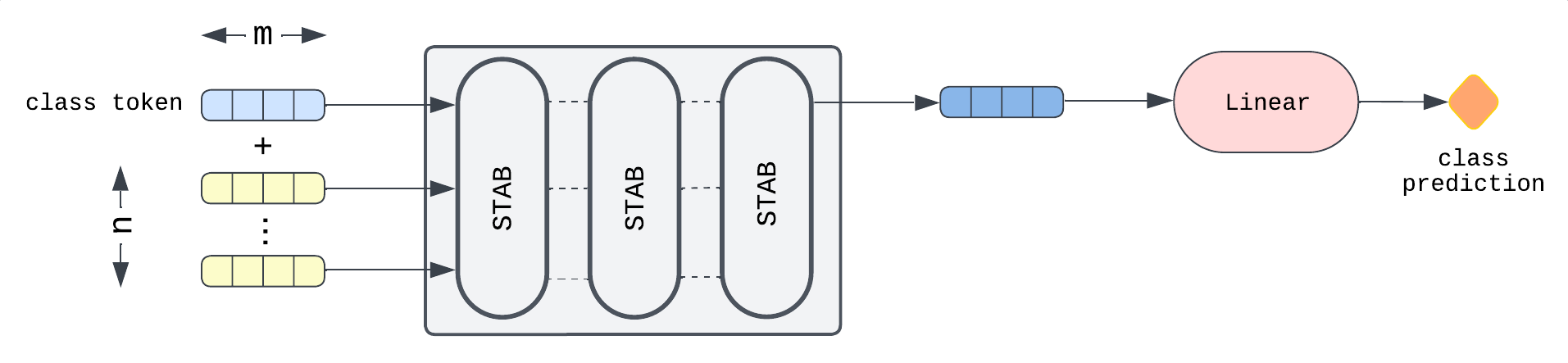}
   \caption{FCM-Former architecture. The input consists of a sample, represented by the event matrix, and augmented with a class token. There is a sequence of three attention blocks, as introduced in \cite{c2}, followed by a linear classification layer which predicts a label for each sample.}
   \label{fig}
\end{figure*}

\section{Experimental setup}

\subsection{Data}

Immunophenotyping for diagnosing childhood acute leukemia typically engaged the use of multiple tubes per sample, each with a different combination of markers.

FCM data are presented in matrix format, where each sample comprises multiple diagnostic tubes. Each tube holds thousands of events, corresponding to feature vectors of individual cells.
Our model utilizes a fixed number of features, and consists of 18 markers and 4 physical properties, as measured by the forward and side scatter of the laser light of the flow cytometer, thereby standardizing the input. Our training fixed-feature list is as follows: FSC-A, FSC-W, FSC-H, SSC-A, CD45, CD71, CD34, CD19, (i)CD79A, (i)CD3, (i)CD22, CD10, CD5, CD7, CD13, CD117, CD33, SY41, LZ, (i)MPO, CD64 and CD65.

FCM-Former thus involves the aggregation of the features present across three datasets, by following the guidelines presented in \cite{c5}, \cite{c24}, with missing values imputed as zeros.
In our work, we ensured that our model training was not biased, by the presence or absence of markers related with lineage-specific markers, which are typically either used or excluded by experts, following the analysis' conclusion of the initial tubes. 

We represent a single sample by a matrix \(E \in \mathbb{R}^{N \times m}\), where $N$ denotes the number of cells (events) in the sample, and $m$ denotes the number of features per cell (which was 22 in our case, as listed above). $N$ is equal to $t \times K$, where $t$ is the number of tubes (typically \(8-13\)) and $K$ the number of cells in every tube (typically \(10^4 - 10^5\), the exact value varies for every tube and every sample). For every index $n \in {1,...,N}$, $E_n \in \mathbb{R}^{m}$ is a quantitative representation of physical and biological properties of every cell. 

\subsection{Datasets}
We evaluate FCM-Former on samples of blood or bone marrow of pediatric patients with B-ALL, T-ALL or AML. The data set consisting of 960 samples was collected at CCRI from 2011 to 2022, with a BD LSR II flow cytometer or BD FACSSymphony A3 and FACSDiva Software (all Becton Dickinson, San Jose, CA).
The samples were stained using a multi-color approach, based on a CD45-Backbone. Markers against lymphoid lineages in each tube allowed defining potential control cells. Immunphenotyping was essentially performed as proposed in \cite{c5}. 
Sampling and research were approved by local Ethics Committees, and informed consent was obtained from patients, their parents, or legal guardians, according to the Declaration of Helsinki. For all samples ground truth information was acquired by manual immunophenotyping assessment, conducted by CCRI experts.

\subsection{Model}

An overview of the FCM-Former architecture is depicted in Figure~1. Our model incorporates an encoder coupled with a linear classification layer. 
We use an ST encoder as presented in Section 2. 
Inspired by Vision-Transformers (ViT)~\cite{c4}, our model is augmented by an additional class token, a learnable feature vector, into the encoder's input. 
At the output of the ST encoder, the trained class token is retrieved and then fed into a linear classification layer.
We treat our problem as a single-label classification, and use a cross-entropy loss for supervised training.
FCM-Former processes a single sample of FCM data in a single forward pass.
Unlike typical transformer-based approaches that incorporate an embedding step, our model is applied directly to FCM samples, specifically bypassing any form of positional embedding. We set the number of induced points to $m=16$, hidden dimension $d=32$ ,and the number of attention heads to 4, for all three layers.
We train our model for $200$ epochs and use an early stop after $50$ epochs if there is no improvement of the accuracy on the validation set.
Throughout all the experiments, we use the cosine-annealing learning rate scheduler with an initial learning rate of $0.001$, lowering to a minimum of $0.0002$ over $10$ iterations for fine-tuning purposes. The Adam optimizer is applied across these experiments while batch processing is not part of our experimental setup. All training processes are executed using an NVIDIA GeForce RTX 3090.
The resulting model is comparatively lightweight with 31,572 parameters.
The accuracy and the ROC-AUC are used as evaluation metrics. 

\section{Results}

Here we present the results of the conducted experiments, evaluated on accuracy and roc-auc metrics.
To ensure the robustness and generalizability of our results, we implemented a 5-fold cross-validation technique. For all experiments, the data are divided into 660 training samples, 100 validation samples, and 200 test samples. The model demonstrates exceptional proficiency in identifying the lineage of Childhood Acute Leukemia, achieving a peak accuracy of $0.965$ and peak roc-auc value $0.9708$ on the test datasets. 
The average accuracy of the model on test datasets across all folds is $0.9408 \pm 0.0217$ and the average roc-auc respectively is $0.9638 \pm 0.0063$.

We additionally experimented with implementing a cross-attention mechanism in our model and trained it accordingly, using as a query $Q$ the learnable vector of the class token, and $K$, $V$ the linear projections of the input set, as in self-attention.
However, the outcomes of this cross-attention mechanism under-performed compared to self-attention, indicating that cross-attention constrains the model's ability to effectively attend to the most relevant parts and relationships within the entire input dataspace, rather than enhancing it.

Furthermore, our model is adaptable to variability across different clinical centers and devices. It facilitates straightforward retraining on new FCM data, with diverse features, highlighting its scalability and potential for integration into clinical routine. 

We identified major causes for misclassification. Cross-lineage marker expression contributed significantly to errors and was the most common cause. Some misclassifications revealed inherent biological complexity, as seen in cases of mixed phenotype acute leukemias (MPAL). Additionally, cases with minimal blast percentages (less than $5\%$) underscored the impact of low cellularity on accurate classification. Poor sample quality and the resulting compromised data quality may also pose challenges to precise classification. These insights highlight the significance of detailed marker analysis and acknowledging biological heterogeneity in improving machine learning models for the classification of leukemia.
 


\section{CONCLUSIONS}

We proposed FCM-Former, a new and automated method for immunophenotyping to diagnose childhood acute leukemia. We trained FCM-Former in a supervised manner and showed that is capable of generalizing to new, unseen data. To the best of our knowledge, FCM-Former is the first attempt to automate the diagnosis of pediatric acute leukemia using FCM data. FCM-Former employs self-attention mechanisms, enabling it to attend to all cells in the sample at once, taking advantage of the whole high-dimension data-space, and avoiding the information loss encountered in the traditional process of manual immunophenotyping assessment. The average performance metrics underscore the FCM-Former's consistent reliability and effectiveness, in diagnosing childhood acute leukemia.

For future work, we would like to extend and improve the performance of our model to predict the mix-lineage and the sub-types of childhood acute leukemia, using only FCM data.



\vspace{12pt}

\end{document}